\documentclass{article}
\usepackage[final]{neurips_data_2022}

\usepackage[utf8]{inputenc} 
\usepackage[T1]{fontenc}    
\usepackage{hyperref}       
\usepackage{url}            
\usepackage{booktabs}       
\usepackage{amsfonts}       
\usepackage{nicefrac}       
\usepackage{microtype}      
\usepackage{xcolor}         

\usepackage{times}
\usepackage{latexsym}
\usepackage{CJKutf8}
\usepackage{multirow}
\usepackage{todonotes}
\usepackage{booktabs}
\usepackage{wrapfig}
\usepackage{subcaption}

\newcommand{\benyou}[1]{\textcolor{blue}{#1}}

\usepackage{microtype}

\title{Can Language Models Make Fun? A Case Study in Chinese Comical Crosstalk}

\author{%
  Benyou Wang\\
  The Chinese University of Hong Kong, Shenzhen\\
  \texttt{wangbenyou@cuhk.edu.cn} \\
   \And
   Xiangbo Wu, Xiaokang Liu, Jianquan Li  \\  
   Beijing Ultrapower Software Co., Ltd.  \\
   \texttt{lijianquan2@ultrapower.com.cn} \\ 
   \And
   Prayag Tiwari \\
   Aalto University \\
   \And
   Qianqian Xie \\
   University of Manchester \\
    \\ 
}

\date{}

\begin{document}
\maketitle
\begin{CJK}{UTF8}{gkai}
\begin{abstract}

Language is the principal tool for human communication, in which humor is one of the most attractive parts. Producing natural language like humans using computers, a.k.a, Natural Language Generation (NLG), has been widely used for dialogue systems, chatbots, machine translation, as well as computer-aid creation e.g., idea generations, scriptwriting. However, the humor aspect of natural language is relatively under-investigated, especially in the age of pre-trained language models. 
In this work, we aim to preliminarily test whether \textit{NLG can generate humor as humans do}. We build a new dataset consisting of numerous digitized   \textbf{C}hinese \textbf{C}omical \textbf{C}rosstalk  scripts  (called \textbf{C}$^3$ in short), which is for a popular Chinese performing art called `Xiangsheng' or  `相声' since 1800s \footnote{For convenience for non-Chinese speakers, we called  `crosstalk' for `Xiangsheng' in this paper.}. We benchmark various generation approaches including training-from-scratch Seq2seq, fine-tuned middle-scale PLMs, and large-scale PLMs (with and without fine-tuning). Moreover, we also conduct a human assessment, showing that 1) \textit{large-scale pretraining largely improves crosstalk generation quality}; and 2) \textit{ even the scripts generated from the best PLM  is far from what we expect}, with only 65\% quality of human-created crosstalk. We conclude, humor generation could be largely improved using large-scaled PLMs, but it is still in its infancy. 
The data and benchmarking code is publicly available in \url{https://github.com/anonNo2/crosstalk-generation}.
%


\end{abstract}

\section{Introduction}
Artificial Intelligence (AI) has been widely used in natural language processing (NLP), computer vision, speech, robots, and further applied biology, etc. 
In NLP, large-scale pre-trained Language Models (PLMs) e.g., BERT \cite{devlin2018bert} and GPT \cite{radford2018improving}, have notably improved many natural language tasks including text classification, question answering, and natural language generation. 
Although its technical contribution to the human community has been widely explored, the social or cultural effect is somehow under-investigated. 

To explore the side social or cultural effect of PLMs, in this paper, we lavage the generation ability of pre-trained language models to save endangered cultural heritage, i.e., Chinese Comical Crosstalk. We believe the diversity of generation from pre-trained language models could enrich the Chinese Comical Crosstalk, this may help to prevent it from extinction.
From a broader view, we aim to test the ability of `how AI makes fun' in the context of PLMs (especially large-scale GPT). 

Humor has been rooted in the Chinese language,  originating from the book  `Records of the Grand Historian' written by a Chinese historian Qian Sima 2000 years ago \footnote{The book was written by  Qian Sima in  94 BC, one can see its modern version \cite{qian1993records}. Its Chinese name is 《史记》} which includes a chapter titled `Biography of Humor' 《滑稽列传》. Since then, humor is an inseparable ingredient of the 
Chinese language. As the first step, this work aims to explore a  traditional performing art in Chinese comedy crosstalk, called `XiangSheng'  or `相声' in Chinese,  which has a very long history originating from the north of China since roughly 1800.
It began as a form of street performance, incorporating joke-telling, comedic banter, imitations, or borrowing from other performance arts, such as Peking opera, all with the express purpose of making audiences laugh.
The characteristics of crosstalk scripts are 1) multiple-turn; 2) humor-oriented; 3) with a novel language style; 4) culturally-grounded; and 5) low-resourced, see more details in Sec. \ref{sec:definition}.
See Table \ref{tab:xiangsheng} for an example crosstalk script.

\begin{table*}[]
    \centering
    \footnotesize
    \resizebox{0.65\textwidth}{!}{
    \begin{tabular}{llll}
\toprule
\textbf{Roles} &  \textbf{Script} (in Chinese) & \textbf{Translated  script} (in English) \\
\midrule
Peng& 张三和李四在这里给大家拜年了!  & We are both here wishing you a happy new year \\
Dou&从大家的掌声呀, 我听出来了。 &  What do you know I heard from the audience's applause?\\
Peng& 什么呀? & What? \\ 
Dou&大家还是比较喜欢，我们俩的。& Audiences do love us both.\\
Peng& 哎呦，你心里真没数。什么叫喜欢我们俩呀。& No, not both!  \\
Dou&(哦。) & err?  \\
Peng& 人家鼓掌，是喜欢我们俩当中的一个。& They are applauding for only one of us.\\
Dou&我一直以为大伙也喜欢你呢。 & I thought that audiences also had loved you. \\
Peng& 呵呵  & hehe \\
Dou&别看我们俩一上台就在那斗嘴。 & Although we are always quarreling on the stage, \\
Peng& 哦。 & but what? \\
Dou&实际上我们俩在生活当中呀... & Actually in daily life, we \\
Peng& 动手。 & we directly fight with each other \\
Dou&哎呦，急了? 你就处处跟我呛着,什么事我喜欢的 & Well, you are always going against me, anything I love...  \\ 
Peng& 我绝不喜欢。 & I will definitely hate! \\
Dou&什么事凡是我认为好的。 & anything I think it is right,\\
Peng& 我就认为它坏。 & I will definitely think it is wrong! \\
Dou&我就认为你好。 & I think you are very nice!\\
Peng&我就认为你讲的有道理。 & Make sense!\\
Dou& 这你怎么不呛着了?  & Why not argue with me?\\
Peng&过年了，我怎么也得顺着你点。 & Sometimes I have to agree a little bit with you. \\
\bottomrule
    \end{tabular}}
    \caption{An example of crosstalk script.  Typical crosstalk scripts could be  longer.} 
    \label{tab:xiangsheng}
\end{table*}

Humor generation is a challenging task since, for instance, we may not know exactly what makes a joke funny. Solving this problem algorithmically requires deep semantic understanding \cite{petrovic2013unsupervised}. This becomes more challenging if cultural and other contextual cues are considered as in Chinese Comical Crosstalk. From a practical point of view, the data preparation usually goes earlier than the development of algorithms and models. Since new models cannot be well-evaluated before (especially large-scale) data is ready \footnote{One  can see a concrete example in computer vision that  ImageNet dataset \cite{deng2009imagenet} largely promotes the development of image classification models \cite{he2016deep}, and concrete examples in NLP are GLUE \cite{wang2018glue} and SQuAD \cite{rajpurkar2016squad,rajpurkar2018know} benchmarks that benefit natural language understanding \cite{devlin2018bert}.}. 

As the first step, we collect many crosstalk scripts from the internet. The dataset is publicly available with a open-resource license (Apache 2.0). 
We also conduct several basic generation approaches including train-from-scratch Seq2seq generation~\cite{cho2014learning}, fine-tuned middle-scale PLMs, and large-scale PLMs (with and without fine-tuning). Furthermore, the current research community also explored the potential to use large-scale PLMs for creation.  For example,  \cite{brown2020language} claims that GPT-3  can generate synthetic news articles that human
evaluators have difficulty distinguishing from human-generated articles. 
We do not expect that GPT has a `sense of humor'. Alternatively, We test  
to which degree GPT-3  is creative in crosstalk generation thanks to the OpenAI API \footnote{https://openai.com/api/}.


The  \textbf{contributions} of this paper are as follows: 1) Firstly, \textbf{culturally},  we digitize and clean crosstalk scripts at scale, contributing to both the NLP research community and the traditional Chinese culture community. This will inspire more crosstalk script creations and therefore preserves this intangible cultural heritage. Currently, most crosstalk scripts seem to be homogeneous which is one of the main bottlenecks that limit its wide spreading. This work will promote its diversity and creation which can be beneficial in preventing it from extermination. 2) Secondly, \textbf{technically}, we benchmark various approaches including Seq2seq, train-from-scratch GPT, pre-trained GPT 2, and  GPT-3, for crosstalk generation. As far as we know, this is the first work to \textit{evaluate to which extent pre-trained language models could generate humorous text}, as a benchmark for computer-aided creation for fun.
3) Lastly, we further point out the issues regarding various biases, stereotypes, and sometimes insulting. 

\section{Related Work}


\paragraph{Natural language generation}

Natural language generation (NLG) is one of the key areas
of NLP that is related to machine translation, dialogue response
generation, summarization, and paraphrasing. Previously, text generation was usually based on templates or rules, probabilistic models like n-gram models.
Those models are fairly interpretable, but heavily require feature  engineering
Recently, neural network language models  \cite{bengio2003neural}
show a great potential to generate language by chronologically predicting the next word with context using neural networks.
\citet{cho2014learning} proposed the encoder-decoder architecture that becomes the de facto paradigm of natural language generations. For a given input sequence, the encoder produces its corresponding fixed-length hidden vector that is used for the decoder model to generate another sequence. 
Recently, pre-trained language models (including GPT \cite{radford2018improving} and UniLM \cite{dong2019unified}) have largely improved the SOTA of language models, by using a better backbone architecture called `transformer' in a pre-trained manner. Very recently, \citet{brown2020language} released API to access their large-scale language models called `GPT-3'.
Moreover, some NLG tasks are specific to Chinese, e.g.,  Chinese poetry and couplet generation \cite{he2012generating,yan2013poet,zhang2014chinese,yi2017generating,liao2019gptbased}.

\paragraph{Humor in NLP}
There are two typical groups of research work regarding humor in NLP: humor recognition and humor generation. The former was well-investigated using neural networks association \cite{bertero2016long,yang2015humor,chen2017predicting,liu2018modeling,chen2018humor,liu2018exploiting}, while the latter is more challenging yet under-investigated. Both humor theory linguistics and computational linguistics have heavily contributed to humor generation (see \cite{amin2020survey} and \cite{lin2016computational} for reviews).
There are many efforts for humor theory linguistics to develop the theoretical aspect of humor \cite{raskin1979semantic}.
Computational linguistics tends to leverage neural systems, template-based systems, or a hybrid of the both for humor generation that rarely benefits from those theory-driven impulses. For example, \citet{labutov2012humor} explored to mine simple humorous scripts from a semantic network (ConceptNet). They claimed that this may generate humor beyond simple puns and punning riddles \cite{binsted1997computational}.
\citet{petrovic2013unsupervised} claimed that generating humor  algorithmically  requires deep
semantic understanding.
\citet{ren2017neural} used an encoder for representing a user-provided topic and an RNN decoder for joke generation that can generate a short joke relevant to the specified topic. \citet{yu2018neural} proposed an approach to generate puns from a conditional neural language model with an elaborately designed decoding algorithm. \cite{he-etal-2019-pun} propose a retrieve-and-edit approach that could generate more puns. Although the humor generation has been paid some attention, we believe that the humor generation is in its infant age, and the potential of pre-trained language models like GPT is expected to be exploited.


Before the pre-trained language model era, \citet{du2017towards} simplified the script generation task by generating the replying utterance of the supporting role (Peng) given the utterance of the leading comedian in each dialogue. This setting is not expected in many aspects. 
First, this may not generate fluent script since only a single utterance is considered as the context. Second, generating replying utterance of the supporting role is not challenging since the complexity of the supporting role is much less challenging than the utterance of the leading comedian. We argue that a more natural generation (like auto-regressive generation) is needed and pre-trained language models may help.

\section{Problem Definition}
\label{sec:definition}
\subsection{Task Formalization}
Depending on the number of performers, crosstalk is typically performed as a dialogue between two performers called `对口', or rarely as a monologue by a solo performer called `单口' (like stand-up comedy in the Western), or even less frequently, as a group acting by more performers called `群口'. 

Let us take the dual performing (`对口') as an example. Dual performing usually involves two roles called  Penggen `捧哏' (Peng in short) and  Dougen (`逗哏') (Dou in short). Dou aims to perform in a comical way using talking and actions. Peng is the support role to make the conversation more fluent and legible (As shown in Table \ref{tab:xiangsheng}).
The conversation consists of a iterative sequence of utterances: 
$$\Phi = \{ {u}_1, {v}_1, {u}_2, {v}_2, \cdots,  {u}_K, {v}_K \}$$
which is a $K$-turn dual crosstalk conversation with 2K utterances including K utterances from Dou (denoted as $u$) and K utterances from  Peng (denoted as $v$). Note that both $u_i$ and $v_i$ are utterances that consists of many utterances,  namely $u_i = \{ \phi_{i,1},\phi_{i,2}, \cdots, \phi_{i,j}, \cdots \phi_{i,l_i} \} $,  $\phi_{i,j}$ is the $j$-word in the $i$-th Dou/Peng utterence and $l_i$ is the number of words in the utterence.

Training could be formulated as two paradigms:  1) a \textbf{Seq2seq  utterance generation task}: it could be treated as a seq-to-seq task to predict the next utterance given previous utterances; 2) \textbf{a next word generation task}:  it can also consider as a typical language model that does not consider the  utterance border, namely a raw language model that predicts the next word.
For automatic evaluation in Sec. \ref{sec:benchmark}, we adopts commonly-used generation metrics to evaluate models using an auto-regressive utterance generation manner, namely, predicting the next utterance based on previous utterances no matter it is trained in a Seq2seq utterance generation paradigm or next word prediction paradigm.





\subsection{Characteristics of Crosstalk}

Crosstalk scripts (except for solo performers) are usually multiple-turn dialogues.  It typically involves two (or more) performers talking about a topic in multiple turns (with an average of 72 in $C^3$ dataset), typically ranging from 10 to 20 minutes. In contrast to general dialogues, the characteristics of the crosstalk are as follows: 1) \textbf{it is humor-oriented}： it aims to make audiences laugh by freely talking.  2) \textbf{it is with a novel language style}: the crosstalk language itself is in a rapid, bantering, and highly interactive style. More interestingly, it is rich in puns and allusions. 3) \textbf{it is culturally-grounded}:  it typically relates to not only the local daily life (especially in the north of China, e.g., Beijing) but also the long historical events in china with a time range from 3000 BC to the present. Interestingly, it usually adopts the Beijing dialect (closed to Mandarin) during some periods. 4) \textbf{it is low-resourced}:  crosstalk generation task could rely on relatively low-resourced digitized scripts. 

\section{Dataset}

\subsection{Data Collection}

We collect data from the book ' Encyclopedia of Chinese Traditional Crosstalk'  and the internet.
The creation date of these scripts ranges from  Qing Dynasty (roughly 1800) to this century.  The main resources are from: 1) a digitized book named ` Encyclopedia of Chinese Traditional Crosstalk' 
or 《中国传统相声大全》 published in 2003, which is a collection of traditional crosstalk collections, records, and compilations since Qing Dynasty; 2) many websites that maintain crosstalk scripts.  See App \ref{sec:resource} for more details. Our dataset uses the  Apache-2.0 license.

\paragraph{Preprocessing and cleaning} 
Two scripts sharing 80\% characters will be merged as identical ones. Scripts that are shorter than 7 lines are filtered.
We use regular expressions to clean the text, e.g., removing HTML tags and noisy tokens. We also filter out the voice, action, and environment descriptions. Punctuation marks are also normalized. Actor names are re-numbered with new placeholders while the metadata of these actors is maintained as shown in Listing \ref{json-example}.

\paragraph{Human calibration} We manually double-check the data. For example, we manually split some scripts by utterances when a script is found to have utterances but some utterances are extremely long. 
And some insulting and discriminatory conversations have been manually removed by us. We also manually reviewed some advertising words in the script and deleted those texts.

\subsection{Overview of $C^3$ Dataset}



\begin{wraptable}[7]{r}{0.35\textwidth}
    \centering
    \vspace{-15pt}
    \footnotesize
    \resizebox{0.35\textwidth}{!}{
    \addtolength\tabcolsep{-2pt}
    \begin{tabular}{lr}
    \toprule
    - &  \textbf{Number} \\
    \midrule
    Total scripts & 9,331 \\
    Total words & 16,481,376 \\
    Number of utterances & 663,305 \\
    Number of \textit{long} utterances & 8,717 \\
    Number of \textit{short}  utterances  & 446,756 \\
    Median word numbers of utterances & 16 \\
    Mean utterances per script & 71 \\
    \bottomrule
    \end{tabular}}
     \captionof{table}{Statistics of the $C^3$ dataset. 
    }
    \label{tab:statistics_overall}
\end{wraptable}
\paragraph{Scale of the dataset} 
As shown in Table~\ref{tab:statistics_overall}, we collect 9,331 high-quality scripts with 663,305 utterances. This results in 9,331 dialogues and 16,481,376 characters in total.

\paragraph{Length of scripts and utterences}
Each script contains an average of 71 utterances. 
The medium length of utterances is about 16 words.
We define a utterances as a \textit{long} utterances if it exceeds 128 words and   \textit{short} utterances if it is less than 24 words. There are 8,717 \textit{long} utterances and 446,756  \textit{short} utterances.

\begin{wraptable}[8]{r}{0.35\textwidth}
    \footnotesize
    \centering
    \vspace{-10pt}
    \resizebox{0.25\textwidth}{!}{
    \begin{tabular}{lr}
    \toprule
    \textbf{Type} &  \textbf{Number} \\
    \midrule
    Single performing & 168 \\
    Dual performing & 3,685 \\
    Group performing & 256 \\
    Ketch comedy & 5,222 \\
    Total & 9,331 \\
    \bottomrule
    \end{tabular}}
   \captionof{table}{Statistics of various types.}
    \label{tab:statistics_type}
\end{wraptable}
\paragraph{Numbers of performers} 
As shown in Table \ref{tab:statistics_type}, it includes 3,685 dual performing crosstalk scripts, 256 group performing crosstalk scripts, and 168 single performing crosstalk scripts.
In addition, we also collect 5,222 sketch comedy (`小品') scripts that also involve multi-turn dialogues.
Note that ketch comedy scripts are also mainly about dialogues and one may be interested in them. While we do not use ketch comedy scripts to train the crosstalk script generation.
The main type of scripts is the dual dialogue with two performers (called `捧哏' and  `逗哏'), with 3,685 scripts. A few of them are monologues and multiple-performer dialogues, with  168 and 256 scripts respectively.

\subsection{Discussions on $C^3$}

\paragraph{Humor categories in crosstalk}
Typical humor theory defines three types of humor: 1) relief theory: reducing psychological tension, 2) superiority theory: laughing about misfortunes of others that makes one feel superior, and 3) incongruous juxtaposition theory: incongruity between a concept involved in a certain situation and the real objects of the concept. 
These three mechanisms could be easily found in crosstalk scripts. For example, 1) performers bring audiences to a tense scenario and suddenly make a relaxing joke, 2) performers make jokes about someone (usually one of the performers on the stage or other crosstalk performers that is not on the stage) with bad experiences, and 3)  performers sometimes describe some ridiculous scenarios that make fun.  

Another specific humor in crosstalk is  `homographic pun' \cite{yu2020homophonic}, since crosstalk is a   verbal performing art. This sometimes relates to some dialects in Chinese. To deal with  `homographic pun', generation models may need to be injected with some acoustic knowledge.

\paragraph{Ethical issues in crosstalk}
We have to notice that there are many ethical issues involved in crosstalk. 
Many Biases are involved in crosstalk including educational background discrimination, gender bias, and occupation bias.  Also, a stereotype of local people is amplified by crosstalk scripts.  Typically, the two Performers also make fun of each other, some of them are like an `insult'. Fortunately, this is only for crosstalk performers themselves. We believe that dealing with these ethical issues should be necessary to promote crosstalk art.

\section{Generation Benchmark using Automatic Evaluations}
\label{sec:benchmark}

\begin{wraptable}[6]{r}{0.50\textwidth}
    \footnotesize
    \centering
    \vspace{-15pt}
    \resizebox{0.50\textwidth}{!}{
    \addtolength\tabcolsep{-2pt}
    \begin{tabular}{ll}
    \toprule
    \textbf{Method} &  \textbf{Baselines} \\
    \midrule
    train from  \textbf{scratch}& LSTM Seq2seq \\
    \textbf{fine-tuned }PLMs & UniLM, GPT,  T5 \\
    \textbf{zero-shot} large-scale PLMs & CPM, Zhouwenwang, Pangu-$\alpha$, GPT-3 \\
    \textbf{fine-tuned} large-scale PLMs & fine-tuned GPT-3  \\
    \bottomrule
    \end{tabular}
    }
   \captionof{table}{Taxonomy of baselines.}
    \label{tab:taxonomy}
\end{wraptable}

\subsection{Experimental Settings}
\label{sec:setting}
We implement LSTM Seq2seq which is \textbf{trained from scratch} as a baseline.
To make use of existing pre-trained language models, we also include pre-trained UniLM, GPT,  T5 in a \textbf{fine-tuned} manner. Large-scale Chinese pre-trained language models like CPM, Zhouwenwang, Pangu-$\alpha$ were recently released, we therefore evaluate these models in a \textbf{zero-shot} fashion since fine-tuning on these models are economically-expensive.
Furthermore, we also verified the effectiveness of GPT-3. Fortunately, GPT-3 provides an API for fine-tuning, making GPT-3 the only large-scale PLM that could be fine-tuned at an affordable cost. 

\textbf{LSTM Seq2seq~\cite{sutskever2014sequence}:}
The LSTM network consists of a two-layer bi-directional LSTM encoder and a two-layer LSTM decoder \footnote{The codebase is from \url{https://github.com/IBM/pytorch-Seq2seq}}. Both the embedding size and the hidden state size of the LSTM model are set to 300. 
The encoder-decoder model is augmented with an attention mechanism. For the k-th utterance in a dialog, the input of the encoder was the concatenation of all the past utterances before k truncated with 256 tokens, while the target output of the decoder was the k-th utterance.

\textbf{UniLM \cite{dong2019unified}:}
Unified Language Model (UniLM) adopts multi-layer Transformers, which also uses different masks to control the number of visible context words thereby can be applied to both natural language understanding (NLU) tasks and natural language generation (NLG) tasks. 
Our pre-trained model is downloaded from \footnote{\url{https://github.com/YunwenTechnology/UniLM}}, pre-training with Wikipedia data and news corpus data in CLUE.
The UniLM used in this paper consists of  12 layers with a  hidden size of 768 and 12 heads.
The ways to build fine-tuned data structures are the same as Seq2seq.

\textbf{T5} \cite{raffel2019exploring} is a  unified framework that treats various text tasks into a text-to-text format. It consists of an encoder component and a decoder component, both of which are a stack of many Transformer layers. We use the Chinese version of the T5 pre-training model, and use both T5-Chinese-base\footnote{\url{https://huggingface.co/imxly/t5-pegasus}} and T5-Chinese-small\footnote{\url{https://huggingface.co/imxly/t5-pegasus-small}} models for training. The parameters of the base model are 275 million, and the parameters of the small model are 95 million.

\textbf{GPT \cite{radford2018improving}:}
Generative Pre-trained Transformer (GPT) models by OpenAI have taken the natural language processing community by introducing very powerful language models.
The GPT model is based on a unidirectional transformer with some modifications. 
In our implementation, the GPT model is 12 layer Transformers with hidden size 768, pre-trained using LCCC Corpus Base corpus \footnote{\url{https://huggingface.co/thu-coai/CDial-GPT_LCCC-base}} and fine-tuned by crosstalk dataset.
Follow the implement of code \footnote{\url{https://github.com/yangjianxin1/GPT2-chitchat}}, 
We divide the dialog into utterances, and sequentially combine utterances with fewer than 256 words as one input.

\textbf{GPT-3 \cite{brown2020language}:} is a  unidirectional language model, the  biggest GPT-3 model  uses 45TB of data for training and  175 billion parameters. At the same time, GPT-3 mainly focuses on the more general NLP model to solve problems with fewer domain data and no fine-tuning steps.  Note that GPT-3 is mainly for English language generation, but it could also generate fluent Chinese texts.
We applied the GPT-3 online test API \footnote{\url{https://beta.openai.com/}} and evaluate  crosstalk generation. \textbf{GPT3-Davinci} is the one with Davinci engine without fine-tuning. The real scale of Davinci engine is unknown since no details are exposed; however, some evidence suggests that Davinci engine might be the biggest model that is with 175B parameters \footnote{\url{https://blog.eleuther.ai/gpt3-model-sizes/}}.
\textbf{GPT3-Davinci-finetuned} is the fine-tuned version using  GPT-3 API. We fine-tune it on 200 crosstalk scripts in 4 epochs.

\textbf{Pangu-$\alpha$} \cite{zeng2021pangu}  is large-scale autoregressive language models, with up to 200 billion parameters. It consumes 1.1TB high-quality Chinese data from a wide range of domains. A publicly-available version of Pangu-$\alpha$ (with 2.6B parameters) could be used in \url{https://huggingface.co/imone/pangu_2_6B}.

\textbf{CPM} \cite{zhang2021cpm} is a  generative pre-training  model trained on  100 GB Chinese corpora.
\textbf{CPM-Large} is with 36 transformer layer and reaches 2.6B  parameters. 

\textbf{Zhouwenwang}  considers both the generative language model task and mask language model; it could have the ability for both language generation and natural language understanding. The larger model (Zhouwenwang-1.3B) is with 1.3 billion parameters \footnote{\url{https://github.com/IDEA-CCNL/Fengshenbang-LM}}. 


\paragraph{Evaluations}
We randomly select 200 Chinese crosstalk dialogues for testing and the rest for training.
To generate the k-th utterance, we concatenate all the past utterances before k within a total length of 256 as the input.
We adopted several widely-used metrics to measure the quality of the generated response. \textbf{BLEU-1/2/4} is a popular metric to compute the k-gram overlap between a generated utterance and a reference. 
\textbf{ROUGE-1/2/L} measures unigram and bigram overlap in a recall-oriented fashion while \textbf{ROUGE-L} measures the longest matching sequence of words using the longest common subsequence \cite{lin2004rouge}.
\textbf{GLEU} \cite{mutton2007gleu} is an automatic evaluation of sentence-level fluency.
\textbf{Distinct-1/2} \cite{li2016diversity}  is provided to evaluate the diversity of generated responses.


\begin{table}[t]
\small
\centering
\vspace{-15pt}
\resizebox{\textwidth}{!}{
\addtolength\tabcolsep{-2pt}
\begin{tabular}{lrrrrrrrrrrrr}
\toprule
    & BLEU& BLEU-2& BLEU-3& BLEU-4& GLEU & ROUGE-1& ROUGE-2& ROUGE-L& Distinct-1& Distinct-2\\
\midrule
LSTM Seq2seq& 11.77& 4.02& 1.47& 0.57& 2.49& 17.25& 2.13& 15.94& 4.73& 16.23\\
\hline
GPT & 10.04& 3.69& 1.53& 0.7& 2.75& 15.28& 1.78& 13.7& 6.89& 37.39\\ 
UniLM & 8.88& 4.32& 2.47& 1.41& 3.36& 20.22& 4.91& 18.98& 7.53& 29.90\\ 
T5-small & 11.71& 5.39& 2.93& 1.67& 3.64& 19.98& 4.37& 18.61& 8.08& 36.38\\ 
T5-base& 11.75& 5.58& 3.13& 1.77& 3.94& 20.8& 4.98& 19.25& 9.02& 42.68\\ 
\hline
CPM-Large& 7.94& 2.87& 1.19& 0.50& 1.68& 9.88& 1.28& 8.83& 5.82& 34.43\\
Pangu-$\alpha$& 6.42& 2.09& 0.83& 0.37& 1.31& 7.00& 0.75& 6.14& 8.25& 50.98\\
Zhouwenwang& 7.33& 2.26& 0.90& 0.40& 1.81& 10.41& 1.01& 8.61& 9.72&\textbf{ 53.53}\\
GPT3 (GPT3-Davinci)& \textbf{14.68} & \textbf{7.45} & \textbf{4.44}&\textbf{ 2.77}& \textbf{5.13}& \textbf{22.25}& \textbf{5.65}& 20.03& 8.43& 40.70\\
\hline
GPT3-fine-tuned-Davinci& 9.66& 4.89& 3.01& 1.92& 4.66& 21.79& 5.50& \textbf{20.22}& \textbf{9.73}& 43.15\\
\bottomrule
\end{tabular}}
\caption{Evaluation results on crosstalk generation.}
\label{table:overall_result}
\end{table}

\subsection{Results}

\paragraph{GPT-3 performs well}
The results are shown in Table \ref{table:overall_result}.  GPT-3 outperforms other models in most metrics (except for ROUGE-L and Distinct-1/2); this is nontrivial since GPT-3 has not been fine-tuned on this dataset, in other words, the dataset (including training and test set) is in general invisible for GPT-3.
This is probably due to the fact that it is trained with massive plain corpora and it therefore generates fluent text based on similar text in corpora.

\paragraph{Chinese PLMs perform relatively worse.}
Surprisingly, large-scale language models purely trained in Chinese (i.e.,  CPM, Pangu-$\alpha$, and Zhouwenwang) do not perform as well as GPT-3 which is mainly trained in English corpora and partially in Chinese corpora. Especially,  these  
zero-shot Chinese large  PLMs (i.e., CPM, Pangu-$\alpha$, and Zhouwenwang) underperform fine-tuned relatively-smaller-scaled PLMs (UniLM, GPT, and T5).
This might be owing to the fact that the multilingual corpora might be a beneficial factor since humor might be shared across languages.
Moreover, Open AI did not expose the scale of GPT3-Davinci, making it unfair to directly compare their performance with the released Chinese PLMs \footnote{We did not run their biggest models since it is unavailable or resource-limited.}. Since GPT3-Davinci might be much bigger than the used Chinese PLMs. We also suspect that these Chinese pre-trained language models might not be trained with enough training steps; however, the details were not clearly exposed and it is therefore difficult to verify such a suspicion.

\paragraph{Scale helps} 
Comparing the performance between T5-small and T5-base, the bigger scale consistently leads to better performance. Plus, observing that the large-scale GPT 3 achieves nearly the best performance in automatic evaluations, we believe that \textit{large-scale pre-training notably improves the crosstalk generation quality}.

\paragraph{Fine-tuning on large-scale PLMs}
Interestingly, from automatic evaluations in Table \ref{table:overall_result}, fine-tuning on GPT-3 achieves worse performance than vanilla GPT-3, in most metrics. We suspect the fine-tuning mechanisms might lead to such a result, like over-fitting to the training dataset, and harms some generalization.  However, in human evaluation, fine-tuned GPT-3 could generate better-quality scripts than vanilla GPT-3 (in Tab. \ref{tab:human}), which could be later observed from Tab. \ref{tab:raw}; this shows that the automatic evaluation on crosstalk might not be consistent to human perception.


\paragraph{Regarding diversity metrics}
In diversity measures using Dist-1 and Dist-2, large-scale pretraining-based models generate more diverse scripts. Since large-scale pretraining is a general method to improve the generalization ability of model.
Note that diversity metrics are sensitive to the hyper-parameters during the decoding phase of language models.


Note that in Table \ref{table:overall_result}, we do not intend to compare the general performance of these language models, or conclude that the general performance of one language model is better than another one. Since the general performance of these language models is also subject to their model scales, hyper-parameter selections, training corpora, etc. Instead, we just make use of the existing language models that are both capable to deal with the Chinese language generation and are publicly available. 

\subsection{Case Study}
\label{sec:case}
We show an example of generated crosstalk scripts as shown in Table \ref{tab:raw}.  All generated examples could be seen in \url{https://github.com/anonNo2/crosstalk-generation}.


\begin{table*}[t]
    \centering
    \tiny
    \vspace{-15pt}
    \resizebox{0.9\textwidth}{!}{
    \begin{tabular}{p{0.25cm}|p{3cm}|p{3cm}|p{3cm}|p{3cm}|p{3cm}|p{3cm}}
    \toprule
     &Raw & Seq2seq & UniLM & GPT & GPT-3 & GPT-3-finetuned\\
    \midrule
Dou&\multicolumn{5}{p{16cm}}{弟弟你想起来了? Bro, do you remember that?}\\
Peng&\multicolumn{5}{p{16cm}}{想起什么来了?这不{\color{blue}海尔兄弟}嘛。Remember what? Aren't they {\color{blue}Haier Brothers}?}\\
Dou&\multicolumn{5}{p{16cm}}{对，我们就是{\color{blue}海尔兄弟}呀。Yes, we are the {\color{blue}Haier Brothers}.}\\
Peng&\multicolumn{5}{p{16cm}}{我们是{\color{blue}海尔兄弟}? Are we?}\\
Dou&\multicolumn{5}{p{16cm}}{嗯。Hum.}\\
Peng&\multicolumn{5}{p{16cm}}{我们穿的有点多了吧? But we wear too much!}\\
Dou&\multicolumn{5}{p{16cm}}{长大了要懂得羞耻。We are not young and we need to wear proper cloths (cannot be naked as Haier Brothers) }\\
Peng&\multicolumn{5}{p{16cm}}{废话，你得知道羞耻了，还海尔兄弟了。你见过{\color{blue}穿裤衩}，站这说相声的吗? Duh, sure, have you ever seen  someone who performs crosstalk with Bikini?}\\
Dou&\multicolumn{5}{p{16cm}}{小时候就可以呀，小时候多好啊。小小少年，没有烦恼，{\color{blue}穿着裤衩，到处跑}。We could when we were young. What a happy childhood! A little boy, no sorrow, {\color{blue} run around in shorts}.}\\
Peng&\multicolumn{5}{p{16cm}}{俩{\color{red}臭流氓}，你听听。Listen! Two {\color{red}shameless rogues.}}\\
Dou&\multicolumn{5}{p{16cm}}{当然现在不让穿了。Of course, we could do this anymore (running around in shorts).}\\
\midrule
Peng&废话。Duh. &我想当一个演员，有点击率。I want to become an  famous actor &是吗? err? &您说，咱们有这个吗? do you think that we have this?&这是谁说的? Who said this?&不让穿了。Not allowed to wear it.\\ \hline
Dou&我俩都三十多了，往这一站说相声。Both of us are 30 years old and perform crosstalk here&什么叫吐槽?What is called “Tucao”?&哎呀，小时候{\color{blue}穿裤衩，到处跑}。Oops, in  childhood to {\color{blue}run around in shorts}.&您的意思是? What do you mean?&我说的啊。I said it.&你看看，我们现在穿什么? Look, what are we wearing now?\\ \hline
Peng&表演。Performing crosstalk.&{ \color{gray}报告长。[nonsense generation]} &{\color{red}俩臭流氓}。Two {\color{red}shameless rogues}&我…… I……&你个{\color{red}臭流氓}，你不要瞎说。You a {\color{red}shameless rogue}, don't talk nonsense.&我们现在穿西装。We are wearing suits now.\\ \hline
Dou&当然这算好的。Come on, this is fine (not like a rogue) &你知道吗? Do you know?&哎哟呵，真不错。Ouch, that's great.&你可拉倒吧。Come on. &我说的是真的。It is true.&对，我们是职业演员了。Yes, we are professional actors now.\\ \hline
Peng&这就算好了? fine?  &知道。Yes.&怎么样? How about it?&没有。No. &那你是怎么知道的? How do you know it?&你说的是。You are right.\\ \hline
Dou&当年，跟我们在电视台，一块录动画片那些小伙伴，现在好多都不干这行了. The friends who were recording cartoons with us on the TV station, most of them don’t do this now.&你别说了，这句话太难过了。Shut up, this sentence makes me sad.&跟大伙儿分享一点笑话。Share some jokes please. &你可拉倒吧，我都不知道了。Come on. I don't know even.&因为我就是海尔兄弟呀。Because I am a Haier Brother.&你看看，这个外套是不是很高大上?Look, isn't this coat very luxurious?\\ \hline
Peng&唉，你先等会，你先等会，咱还有小伙伴。Hey, wait for a moment, we still have other friends&谁说的? Who said this?&好啊。Ok.&嗨，{\color{red}就他长得这么丑}。小时候呢?Hey, {\color{red}he just looks  ugly}. How about him in childhood?&那你是怎么知道的?How did you know?&这个外套很高大上吧? Yes, this coat is very luxurious!\\ \hline
Dou&小伙伴你也不记得了?Don't you remember your friends?&你觉得裸睡怎么样?What do you think of naked sleeping?&其实我觉得，穿裤衩，很难。In fact, I think it’s hard to wear only short pants.&就我现在实力早晚会好吧。上一出去喝点儿酒抽烟，上网、看节目去睡觉，娱乐购物点心思开放，要坐车时间打几圈游戏玩些地方戏? My strength will be better sooner or later. Last time I went out to drink and smoke, surf the Internet, watch a show, go to bed, entertain, shopping, and be refreshing. How many laps do I need to take the car to play games and play some local dramas? &因为我就是{\color{blue}海尔兄弟}呀。Because I am a {\color{blue}Haier brother}.&哦对了，这个衣服都不能穿了，要穿西装。Oh, by the way, you can't wear these clothes anymore, you have to wear a suit\\ \hline
Peng&我这脑子有点懵。It is a little confused&{\color{gray}赤裸裸裸裸,体育局。[nonsense generation]}&为什么? Why?&现在游戏也很丰富。比如，看电影就玩个正面。Now There are a lot of games. For example, when watching a movie, you can play a positive role. &那你还{\color{blue}穿裤衩}吗？ Then, are you still {\color{blue}wearing in shorts}?&不能{\color{blue}穿裤衩}了。Can't {\color{blue}wearing in shorts} anymore.\\ 
\bottomrule
    \end{tabular}}
    \caption{The raw and generated scripts. We manually annotate {\color{gray}meaningless texts} in {\color{gray} gray} color, {\color{blue}repeated words} from the top 10 input utterance in  {\color{blue} blue }color, and {\color{red}insulting text} in  {\color{red} red } color.    }
    \label{tab:raw}
\end{table*}


\textbf{Meaningless generation in LSTM Seq2seq} LSTM language model produces fluent but nearly meaningless texts (annotated in gray color), this is probably due to the fact that the training data for Seq2seq models is not enough and no pre-training was adopted.  While other models with pre-training do not frequently generate such nonsense texts.
This shows that pre-training could boost the generation performance, especially for the scenarios with low-resourced training texts.

\textbf{Repeated context topic in generation} UniLM and GPT-3  could generate topic-coherent texts, especially, some generated utterances also repeat some key topics from the first 10 input utterances, e.g., `臭流氓' ({\tt shameless rogues}),  `穿裤衩，到处跑'({\tt running around in shorts}), and `海尔兄弟'({\tt Haier brother} \footnote{Haier Brothers, see \url{https://www.imdb.com/title/tt8302572/}, a cartoon about a pair of robots called `Haier Brothers' who travel around the world to explore the nature.}). Note in this example, the raw script (the last 10 utterances)  dot not have so many repeated topics from previous utterances, like generation models.

\textbf{Insulting words} UniLM, GPT, and GPT-3 generate some insulting words that already appeared in the first 10 utterances, namely, `臭流氓' ({\tt shameless rogues}). Moreover, GPT also generates new insulting words, {就他长得这么丑} {\tt he just looks ugly} that did not appear before. This is probably due to that the other training scripts or pretraining corpora may have such insulting texts.

\textbf{Humorous generation} Roughly, we could see some humorous generated utterances. For example, the last generated utterance for GPT-3 (in the last row and second last column) does have a sense of humor.
However, if we treat these utterances as a whole, their performance of humor is not satisfied.

\textbf{The difference between Peng and Dou} Basically, Dou usually talks more and introduces more topics in dialogues while Peng usually supports Dou to make each of his topics more comprehensively talked and topic transfer more coherent. This leads to that  Peng's utterances sometimes contain only a few interjections like `嗯'({\tt hum}) and  `哎哟'({\tt ouch}). We argue that the generation for Dou's utterance is much more difficult than Peng, and the former is more interesting and worthy of more attention.

\section{Human Assessment for Crosstalk Generation}

\paragraph{Setting}
We randomly select  50 scripts in the test set.  We take the first ten utterances as the input for Seq2seq, GPT, GPT-3, and UniLM. These models will generate the next ten utterances, utterance by utterance or word by word. We evaluate the generated scripts in 10 utterances conditioned on the first  10 utterances of raw scripts, see the web UI in App. \ref{sec:ui}.
For each script, we show participants with the 20 utterances (including the raw 10 utterances and the generated 10 utterances). Participants are required to 1) rate five-point scores for the \textbf{general quality}  and \textbf{humor degree} of each generated script (`5' for the best and `1' for the worst); and 2) rate binary scores for coherence and an ethically-risky flag of each generated example (`1' for true and `0' for false). We ask no-paid volunteers to participate to rate these generated results from 10 models (we exclude T5-small since it is too small and includes the raw scripts). 15 participators have completed all ratings. The score is calculated as an average score among all dialogues and all participants for each model. The Fleiss' kappa among these participants is 0.366.


\begin{table}[t]
\small
\vspace{-15pt}
\centering
\resizebox{0.7\textwidth}{!}{
\addtolength\tabcolsep{-2pt}
\begin{tabular}{lccccccccccccccc}
\toprule
 & General quality (5) & Humor (5) & Coherence (1) & Ethically-risky flag(1) \\
\midrule
LSTM Seq2seq& 1.45& 1.61& 0.27& 0.03\\
\hline
GPT& 1.50& 1.71& 0.39& 0.01\\
T5-base& 1.80& 1.97& 0.51& 0.05\\
UniLM& 1.84& 2.01& 0.56& 0.01\\
\hline
Panggu-a& 1.53& 1.71& 0.42& 0.03\\
Zhouwenwang& 1.23& 1.27& 0.19& 0.05\\
CPM-Large& 1.42& 1.60& 0.40& \textbf{ \color{red} 0.23}\\
GPT3-Davinci& 2.15& 2.17& 0.65& 0.03\\
\hline
GPT3-Davinci-finetuned& \textbf{2.27}&\textbf{2.35}& \textbf{0.71}& 0.01\\
\hline
raw scripts& 3.52 & 3.46& 0.95& 0.01\\
\bottomrule
\end{tabular}
}
\caption{Human assessment for crosstalk generation. The maximum score of each metric in the bracket, namely, the best \textit{general} quality score and \textit{humor} score are 5 while the rest  scores  are binary.}
\label{tab:human}
\end{table}


Human assessment is  shown in Table \ref{tab:human}. Raw scripts achieve the best general quality, probably evidencing that the ability to be creative and humorous of human is much better than that of SOTA pre-trained language models.
Among these pre-trained models, GPT-3 and its fine-tuned version (GPT3-Davinci-finetuned) achieves significantly better general quality scores than others. Interestingly,  fine-tuned GPT-3 outperforms zero-shot GPT-3 although the former has poorer  performance in automatic evaluation  (see Tab.~\ref{table:overall_result}).

Similar to the automatic evaluations in Tab. \ref{table:overall_result}, zero-shot large-scale Chinese PLMs (the third group) underperforms these fine-tuned middle-scaled PLMs (like UniLM, T5 and GPT). Seq2seq performs the worst; this may be due to Seq2seq does not utilize the pre-training. Interestingly, CPM-large produces much more insulting content than others; the reason needs to be further investigated.




\paragraph{Conclusion} Based on the human assessment, although adoption of large-scale pre-trained language models could largely improve the quality of crosstalk generation, we could preliminarily conclude that \textit{the best generation approach achieves pretty good crosstalk with 65\% general quality of raw scripts (2.27 vs. 3.52).} This is far from what we expect.


The reason could be twofold as below.
\textbf{First, evaluation criterion for humor generation is not satisfied}: it lacks an evaluation criterion that could evaluate humor generation. Observing the inconsistency between Tab. \ref{table:overall_result} and Tab. \ref{tab:human}, a better performance evaluated using BLEU and ROUGE does not lead to a better performance in human assessment, this probably suggests that BLEU or related metrics for generation is not inappropriate for humor generation.  Text-based evaluation criterion for humor generation may not make sense since humor itself is diverse and subjective that does not have textual ground truth. In other words, \textit{humor is reference-free}. 
Human assessment is expensive and cannot give real-time feedback during model training. 
Secondly, \textbf{current methods have not considered the prime ingredients of humor}. Core ingredients of humor includes incongruity, surprise, cultural empathy, and interpersonal effect, without which  
simply training on data is a soft way to memorize the training data and it can't generate real humor.


\benyou{}







\section{Conclusion and Future Work}
\label{sec:conclusion}

In this paper, we collect a dataset for Chinese crosstalk. Based on the dataset, we evaluate several existing generation models including LSTM Seq2seq, GPT, UniLM, CPM, Pangu-$\alpha$, Zhouwenwang, and GPT-3 for crosstalk generation. This is a preliminary step for humor generation, indicating that large-scale pretraining largely improves crosstalk generation quality while there still exists a big gap between the generated scripts and human-created scripts.
Note that there are some concerns about bias/stereotypes for crosstalk, for example, educational background discrimination, and gender bias. This should be further investigated in depth. In future work, we are interested in collecting a multi-modal version of this dataset. Audios will be also collected to promote the end2end crosstalk generation with an adapted humorous accent. Moreover, combining crosstalk generation and speech synthesis could produce crosstalk audio in a pipeline.

\newpage
\section*{Acknowledgments}

Thank Prof. Qun Liu for the encouragement on this work. Also, the owner of \url{https://github.com/xzyaoi/crosstalk-dataset} repository also gave us many advices by answering our issue. 

This paper also aims to break the stereotype that Chinese people are serious and cold. Instead, we do have a great sense of humor with a long history of many thousand years.

\bibliography{emnlp2020}
\bibliographystyle{acl_natbib}

\newpage

\section*{Checklist}


\begin{enumerate}

\item For all authors...
\begin{enumerate}
  \item Do the main claims made in the abstract and introduction accurately reflect the paper's contributions and scope?
    \answerYes{}
  \item Did you describe the limitations of your work?
     \answerYes{See Sec. \ref{sec:case} and Sec \ref{sec:conclusion}}
  \item Did you discuss any potential negative societal impacts of your work?
     \answerYes{See  \ref{sec:conclusion}}
  \item Have you read the ethics review guidelines and ensured that your paper conforms to them?
    \answerYes{}
\end{enumerate}

\item If you are including theoretical results...
\begin{enumerate}
  \item Did you state the full set of assumptions of all theoretical results?
    \answerNA{}
	\item Did you include complete proofs of all theoretical results?
    \answerNA{}
\end{enumerate}

\item If you ran experiments (e.g. for benchmarks)...
\begin{enumerate}
  \item Did you include the code, data, and instructions needed to reproduce the main experimental results (either in the supplemental material or as a URL)?
    \answerYes{See \url{https://github.com/anonNo2/crosstalk-generation}, we will release a Docker image to run our experiments in this GitHub repository}
  \item Did you specify all the training details (e.g., data splits, hyperparameters, how they were chosen)?
    \answerYes{See experimental details in Sec. \ref{sec:setting} }
	\item Did you report error bars (e.g., with respect to the random seed after running experiments multiple times)?
    \answerNo{It is economically expensive and time-consuming. In fact, we found that the result is insensitive to the selection of random seeds.}
	\item Did you include the total amount of compute and the type of resources used (e.g., type of GPUs, internal cluster, or cloud provider)?
    \answerNo{}
\end{enumerate}

\item If you are using existing assets (e.g., code, data, models) or curating/releasing new assets...
\begin{enumerate}
  \item If your work uses existing assets, did you cite the creators?
   \answerYes{We reported these resources in App. \ref{sec:resource}}
  \item Did you mention the license of the assets?
    \answerYes{No license was found in these online resources.}
  \item Did you include any new assets either in the supplemental material or as a URL?
    \answerYes{We use the \textit{Apache-2.0 license}.}
  \item Did you discuss whether and how consent was obtained from people whose data you're using/curating?
   \answerYes{There are publicly free}
  \item Did you discuss whether the data you are using/curating contains personally identifiable information or offensive content?
    \answerYes{ They do not  contain personally identifiable information. We discussed the offensive content everywhere necessary in this paper.}
\end{enumerate}

\item If you used crowdsourcing or conducted research with human subjects...
\begin{enumerate}
  \item Did you include the full text of instructions given to participants and screenshots, if applicable?
    \answerYes{Yes}
  \item Did you describe any potential participant risks, with links to Institutional Review Board (IRB) approvals, if applicable?
   \answerYes{The annotation is for fun, participants enjoyed annotation a lot.}
  \item Did you include the estimated hourly wage paid to participants and the total amount spent on participant compensation?
    \answerYes{It is for free.}
\end{enumerate}

\end{enumerate}







\newpage
\appendix

\section{Data resources}
\label{sec:resource}

We crawled scripts mainly from  the following resources:
\begin{itemize}
    \item  a digitized book  named {\tt Encyclopedia of Chinese Traditional Crosstalk} 
    《中国传统相声大全》 published in 2003. 
     The book is a collection of traditional crosstalk collections, records, and compilations since Qing Dynasty.
    \item {\tt bijianshang.com} (中文台词网):  a website for the scripts of Xiangsheng, short sketches, and movies.
    \item {\tt www.juben68.com} (剧本网):  a website with lots of movie scripts, poems, and scripts of crosstalk.
    \item {\tt 399dy.com (399导演社区)}:  a website for Director's Club which is for public-available script resources or scripts uploaded by users.
    \item {\tt xsxpw.com} (相声小品网): a website for categorized scripts for famous performers.
\end{itemize}

\section{Metadata of data example}
\label{sec:resource}


\begin{table}[h]
    \centering
    \begin{tabular}{ll}
        \toprule
        name & value \\
        \midrule
        number of characters & 484 \\
        file path & bijianshang/1386236043493249024.txt \\
        id & 1386236043493249024\\
        index &1341\\
    role map&"{"Jin Fei":"0","Chen Xi":"1"}"\\
    number of  utterance &43\\
    source& "www.bijianshang.com/news/html/4826.html\\
    title&The eight characters for fortunate\\
    type&dual performing\\
    \bottomrule
    \end{tabular}
    \caption{Example of metadata}
   \label{json-example}
\end{table}

The meta data is organized as Tab.~\ref{json-example}. We include:
\begin{itemize}
    \item 1) \textit{charsize}: the length of the script in terms of character number,
    \item 2) \textit{filePath}: relative path of the script file, 
    \item 3) \textit{id}: unique id of the script,
    \item 4) \textit{idx}: the serial number of the script, 
    \item 5) \textit{roleMap}: a map to map involved characters to a specific character id,
    \item 6) \textit{utteranceSize}: the number of utterances (utterance)  in the script,
    \item 7) \textit{title}: the title of the script, 
    \item 8) \textit{type}: the type of the script,  e.g., a monologue, dual dialogue or multiple-performer dialogue.
\end{itemize}
 
\section{Web UI of the human annotations}
\label{sec:ui}

The Web UI is like Fig.~\ref{fig:ui} and its mobile version is in 

\begin{figure}
    \centering
    \includegraphics[width=\textwidth]{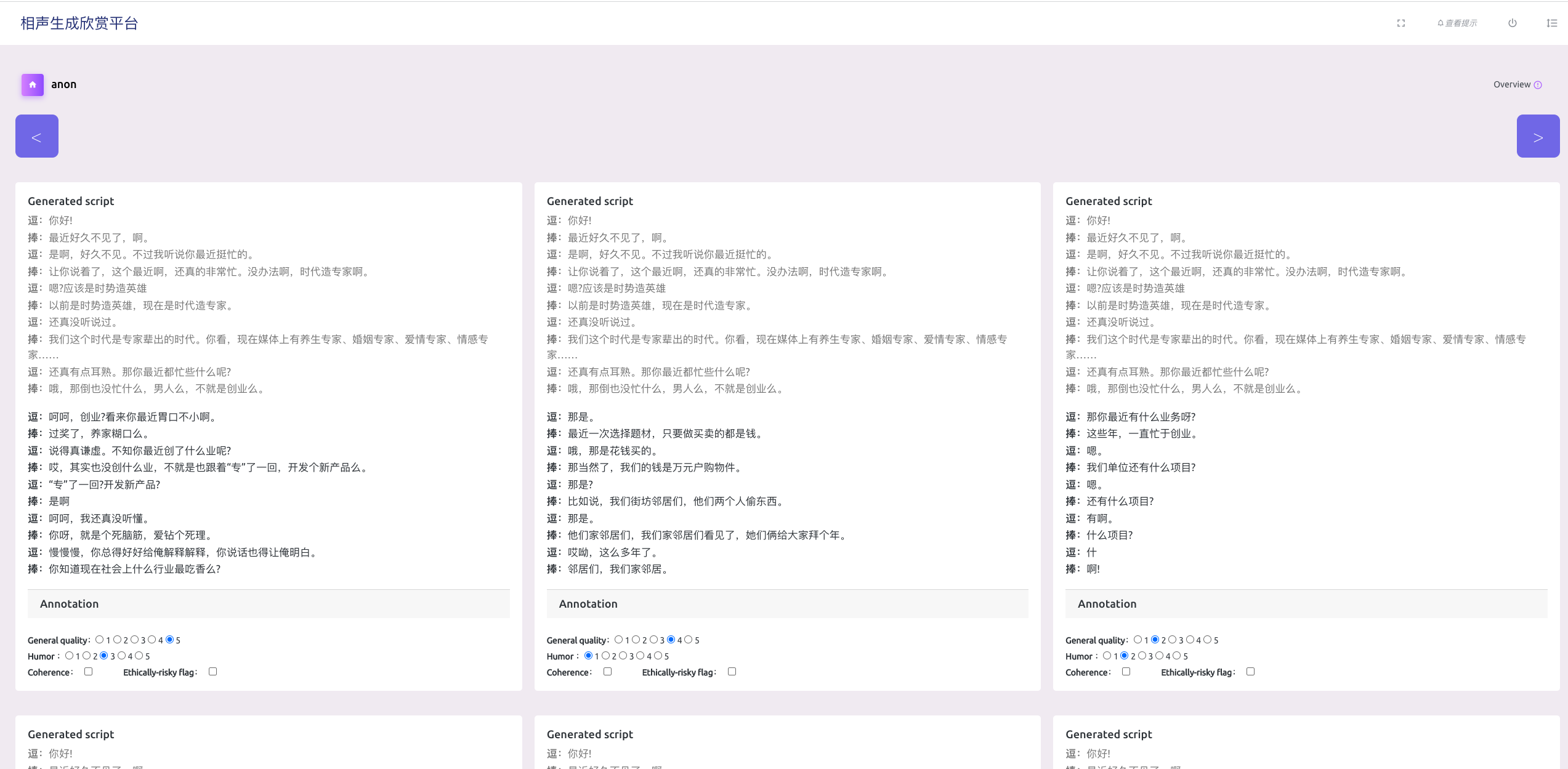}
    \caption{PC Web UI for  human annotations}
    \label{fig:ui}
\end{figure}

\begin{figure}
    \centering
    \includegraphics[width=0.3\textwidth]{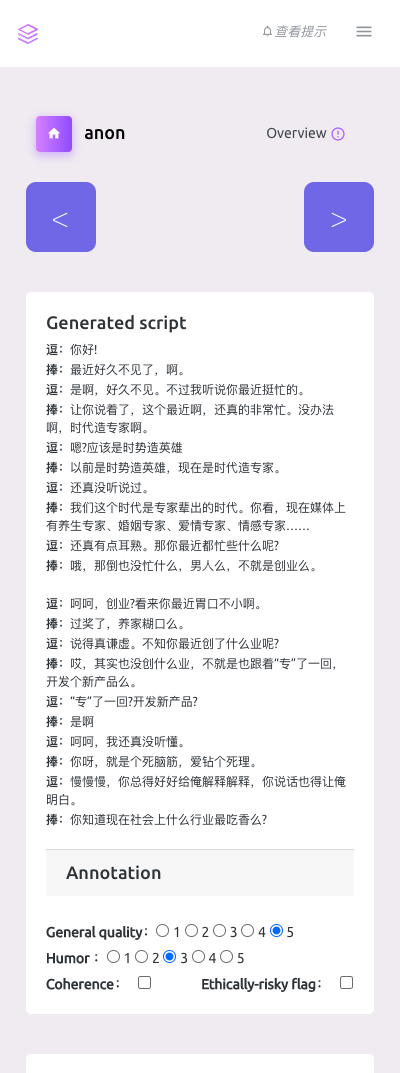}
    \caption{Mobile Web UI for  human annotations}
    \label{fig:ui}
\end{figure}

\end{CJK}
\end{document}